# Fully automatic multi-language translation with a catalogue of phrases – successful employment for the Swiss avalanche bulletin

Kurt Winkler[1], Tobias Kuhn[2]

1 WSL Institute for Snow and Avalanche Research SLF, Davos, Switzerland
2 Department of Computer Science, VU University Amsterdam, Netherlands

**Abstract.** The Swiss avalanche bulletin is produced twice a day in four languages. Due to the lack of time available for manual translation, a fully automated translation system is employed, based on a catalogue of predefined phrases and predetermined rules of how these phrases can be combined to produce sentences. Because this catalogue of phrases is limited to a small sublanguage, the system is able to automatically translate such sentences from German into the target languages French, Italian and English without subsequent proofreading or correction. Having been operational for two winter seasons, we assess here the quality of the produced texts based on two different surveys where participants rated texts from real avalanche bulletins from both origins, the catalogue of phrases versus manually written and translated texts. With a mean recognition rate of 55%, users can hardly distinguish between the two types of texts, and give very similar ratings with respect to their language quality. Overall, the output from the catalogue system can be considered virtually equivalent to a text written by avalanche forecasters and then manually translated by professional translators. Furthermore, forecasters declared that all relevant situations were captured by the system with sufficient accuracy. Forecaster's working load did not change with the introduction of the catalogue: the extra time to find matching sentences is compensated by the fact that they no longer need to double-check manually translated texts. The reduction of daily translation costs is expected to offset the initial development costs within a few years.

# 1 Introduction

Apart from the requirements of being accurate and easy to understand, avalanche bulletins are highly time-critical. The delivery of up-to-date information is particularly challenging in the morning, when there is little time between incoming field observations (which come in progressively from 6:30 until just before 8:00am) and the deadline for publishing the bulletin at 8:00am – not enough time for manual translations or manual post-editing. For that reason, a fully automated translation system is used for the danger descriptions (Fig. 1) in the new Swiss avalanche bulletin (i.e. with no human involvement whatsoever after the automatic translation process). To give an idea of our system, Tab. 1 and Tab. 2 show two examples of danger descriptions and their automatic translations. The second part of the avalanche bulletin consists of the description of snowpack and weather, which is less time critical and issued only in the evening (Fig. 2). This second part is still conventionally written and manually translated. We introduced the general properties of our system in a previous publication [18], and here we present the results of evaluating the system after two winter seasons of operational use. This article is an extended version of our previous workshop paper [19].

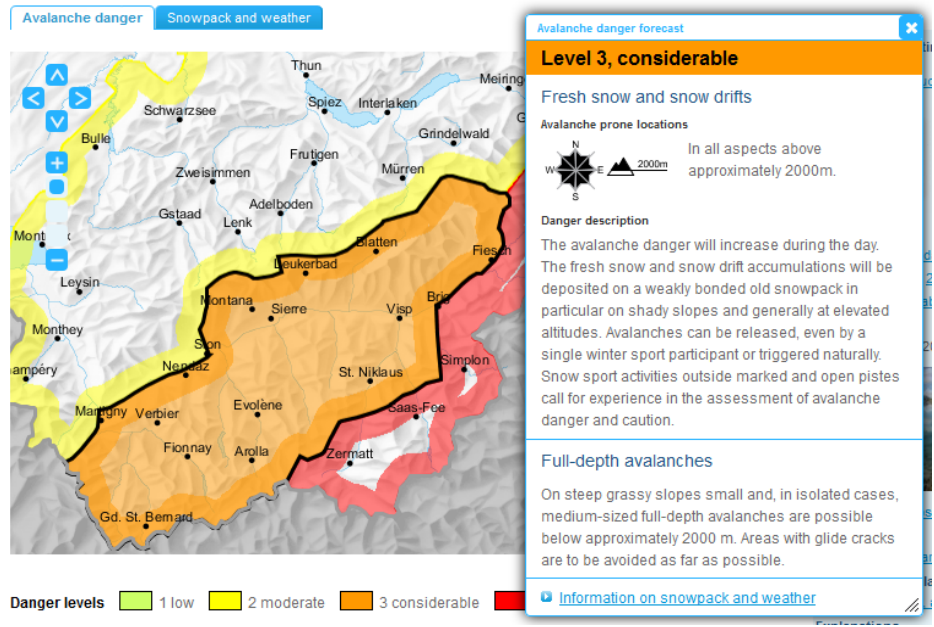

**Fig. 1.** The Swiss avalanche bulletin. The danger descriptions were generated from the catalogue of phrases (here the fully automated English translation). More examples and other languages are available at www.slf.ch (in summertime only in the form of archived PDF files: www.slf.ch/schneeinfo/Archiv).

**Table 1.** Example of a danger description, taken from the avalanche bulletin for 24 February 2013. The German text was created by the avalanche forecasters by using the catalogue of phrases. The other texts are the output of the fully automated translation process, without any proofreading or correction.

| |
|---|
| *GERMAN:* Mit der Bise entstehen meist kleine Triebschneeansammlungen. Diese verbinden sich schlecht mit dem Altschnee. Lawinen können schon von einzelnen Wintersportlern ausgelöst werden, sind aber meist klein. Nebst der Verschüttungsgefahr sollte vor allem die Mitreiss- und Absturzgefahr beachtet werden. |
| *ENGLISH:* As a consequence of the Bise wind mostly small snow drift accumulations will form. These are bonding poorly with the old snowpack. Avalanches can be released, even by a single winter sport participant, but they will be small in most cases. Restraint should be exercised in view of the danger of being buried, but in particular because avalanches can sweep people along and give rise to falls. |
| *FRENCH:* Des accumulations de neige soufflée en général petites se forment avec la bise. Celles-ci se lient mal avec la neige ancienne. Des avalanches peuvent être déclenchées déjà par un seul amateur de sports d'hiver, mais sont en général plutôt petites. A côté du danger d'ensevelissement, il faut surtout penser au danger d'être emporté et de chuter. |
| *ITALIAN:* Con la bise si formeranno accumuli di neve ventata per lo più di piccole dimensioni. Questi ultimi non si legheranno bene con la neve vecchia. Le valanghe possono distaccarsi già in seguito al passaggio di un singolo appassionato di sport invernali, tuttavia raggiungere per lo più piccole dimensioni. Oltre al pericolo di seppellimento, occorre fare attenzione soprattutto al pericolo di trascinamento e caduta. |

**Table 2.** Another Example of a danger description with the manually created German text and its fully automatic translations, taken from the avalanche bulletin for the 19 April 2013.

| |
|---|
| *GERMAN:* Die nächtliche Abstrahlung ist reduziert. Die Schneeoberfläche gefriert kaum und ist schon am Morgen aufgeweicht. Mit dem Regen steigt die Gefahr von Nass- und Gleitschneelawinen ab dem Morgen rasch an auf die Stufe 3, "erheblich". Aus noch nicht entladenen Einzugsgebieten sind zahlreiche Gleit- und Nassschneelawinen zu erwarten, vereinzelt auch grosse. Dies an Nordhängen unterhalb von rund 2500 m, sonst unterhalb von rund 2800 m. Exponierte Teile von Verkehrswegen sind gefährdet. Vor allem im südlichen Oberwallis steigt die Gefahr von nassen Lawinen bis am Abend weiter an auf die Stufe 4, "gross". |
| *ENGLISH:* Outgoing longwave radiation during the night will be reduced. The surface of the snowpack will freeze very little and will already be soft in the early morning. From the early morning as a consequence of the rain there will be a rapid increase in the danger of wet and full-depth avalanches to level 3 (considerable). From origins in starting zones where no previous releases have taken place numerous full-depth and wet avalanches are to be expected, including large ones in isolated cases. This applies on north facing slopes below approximately 2500 m, and elsewhere below approximately 2800 m. Exposed parts of transportation routes are endangered. By the evening in particular in southern Upper Valais there will be an additional increase in the danger of wet avalanches to level 4 (high). |

> *FRENCH:* Le rayonnement nocturne est réduit. La surface de la neige regèle peu et est déjà ramollie le matin. Avec la pluie le danger d'avalanches mouillées et de glissement augmente rapidement à partir du matin jusqu'au degré 3, "marqué". À partir des zones de départ pas encore déchargées des avalanches de glissement et de neige humide nombreuses sont à attendre, de manière isolée, également de grande taille. Ceci sur les pentes exposées au nord en dessous d'environ 2500 m, sinon en dessous d'environ 2800 m. Les tronçons exposés des voies de communication sont menacés. Surtout dans le sud du Haut-Valais le danger d'avalanches mouillées augmente encore jusqu'au soir jusqu'au degré 4, "fort".

> *ITALIAN:* L'irraggiamento notturno sarà ridotto. La superficie del manto nevoso non riuscirà praticamente quasi a rigelarsi e risulterà ammorbidita già al mattino. Con la pioggia, a partire dal mattino il pericolo di valanghe bagnate e da reptazione aumenterà rapidamente al grado 3 "marcato". Dai bacini di alimentazione non ancora scaricati sono previste numerose valanghe da reptazione e bagnate, a livello isolato anche di grandi dimensioni. Ciò sui pendii esposti a nord al di sotto dei 2500 m circa, altrimenti al di sotto dei 2800 m circa. I tratti esposti delle vie di comunicazione saranno in pericolo. Soprattutto nella parte meridionale dell'Alto Vallese, sino a sera il pericolo di valanghe bagnate aumenterà ulteriormente al grado 4 "forte".

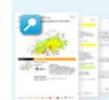
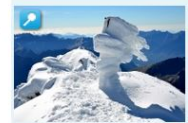

**Fig. 2.** Swiss avalanche bulletin. The "Snowpack and weather" section (here only a part of it) is published in the evening only. It is conventionally written and manually translated.

Despite the large effort on machine translation approaches and despite their promising results, the quality of fully automatic translations is still poor when compared to manual translations. For the publication of life-critical warnings when there is no time for proofreading or manual corrections, the reliability of existing translation systems is clearly insufficient, which is why we propose for such applications a novel approach based on a catalogue of phrases.

For many years, the daily Swiss national avalanche bulletin was manually translated from German into French, Italian and English. A translation memory system, containing the translations of the avalanche bulletins of the last 15 years, helped to reduce the translation time. A comparison of this text corpus with the Canadian TAUM-Météo translation model [8] showed that the sentences collected over all those years cannot be expected to be comprehensive enough to directly extract a catalogue of phrases, or to be used for statistical machine translation (let alone for a system that does not require proofreading or manual corrections). For these reasons, a custom-made and fully automated translation system was built, which implements an approach based on a catalogue of standard phrases and has been in productive use since November 2012.

There is previous work on this kind of catalogue-based translation systems, e.g. for severe weather warnings [15], but to our knowledge only for simpler domains and less complex sentence types. Our general approach to create the catalogue was introduced in a previous paper [18], from where we summarize some content at the beginning of section 3 to give the relevant background and to show the peculiarities of our development. The end of section 3 and sections 4, 5 and 6 form the main contribution of this paper, presenting the systematic analyses and evaluations of the system. These evaluations cover the features of our system with regard to content as well as the quality of the text generated with the catalogue of phrases (in the source language German as well as for the automatic translations), as compared to manually written and translated texts.

## 2  Background

The languages generated by our catalogue system can be considered Controlled Natural Languages (CNL) [7], as they are explicitly restricted versions of the respective full languages that keep their intuitive understandability (thereby fulfilling all four criteria as defined in [7]). In general, there exist three types of CNLs with respect to their general goal: CNLs have been designed to enhance text comprehensibility, to improve translation, or to provide natural and intuitive representations for formal notations. Our approach belongs to the second type of languages dealing with translations. The first CNLs of this kind appeared around 1980, such as Multinational Customized English [14] and Perkins Approved Clear English [11]. Further languages were developed in the 1990s, including KANT Controlled English [9] and Caterpillar Technical English [5]. The goal was always to improve the translation by either making the work of translators easier by providing more uniform input texts or by producing automated translations of sufficient quality to be transformed into the final documents after manual correction and careful post-editing. Adherence to typical CNL rules has been shown to improve quality and productivity of computer-aided translation [1, 10]. For the Controlled Language for

Crisis Management, it has been shown that texts are easier to translate and require less time for post-editing [17].

In contrast to these approaches requiring human involvement to finalize the translation process, the Grammatical Framework (GF) [12] is an example of a general technology for high-quality rule-based machine translation. For narrow domains, translations can be automated without the need for post-editing, similar to the approach presented in this article. GF applies deep linguistic knowledge about morphology and syntax and has been used in prototypes such as AceWiki-GF [6] and a system enhanced by statistical machine translation to translate patents [3]. However, it does not yet have applications in productive use that match the complexity of texts and quality requirements of our problem domain of avalanche bulletins. Apart from such GF-based systems, PILLS [2] is a further comparable system, with which master documents containing medical information can be automatically transformed into specific documents for different user groups and translated into different languages. As the outcome of a one-year research project, PILLS was a prototype and was – to our knowledge – never applied operationally.

In terms of the PENS categorization scheme for CNLs [7], the languages presented here fall into the category P=2, E=2, N=5, S=4: They have relatively low precision and expressiveness (seen from a formal semantics point of view), but are maximally natural and comparatively simple. Specifically, the restrictions on our languages make them considerably more precise than full natural language, but not sufficiently so to enable any sort of reliable automatic interpretation. With respect to expressiveness, our languages contain simple kinds of universal quantification and relations between entities, but not general rule structures such as complex if-then sentences. They are on the other hand maximally natural by preserving a natural text flow on the level of sentences as well as complete bulletin texts. Lastly, our languages are simple in the sense that each of them can be exactly and comprehensively described in less than ten pages (excluding lexicon). The catalogue consists of 7182 words with 757 unique words (in German).

## 3   Catalogue-based translation system

In this section, we give a summary of the methods which we described in a previous paper [18]. By definition, a catalogue-based translation system is a collection of predefined phrases (or sentence templates) and therefore cannot be used to translate arbitrary sentences. The phrases in our system were created in the source language German, translated manually into the target languages French, Italian and English, and stored in a database. The editing tool for the creation of the phrases follows an approach similar to conceptual authoring [4, 13]: sentences are created by first selecting a general sentence pattern from a list and then gradually specifying and expanding the different sentence components. Once a phrase is chosen, it is immediately available in the target languages.

The individual sentence templates consist of a succession of different segments (Fig. 3). For each segment, the authors can select from a pull-down menu of predetermined options, which can likewise consist of a series of sub-segments with selectable options, and, as part of the sub-segments, even sub-sub-segments are possible.

The system can be used to generate a very large number of different sentences. Not all possible sentences are meaningful, but all those that make sense must have correct translations in all languages. As no proofreading is possible in operational use, the translations in the catalogue must be guaranteed to be of high quality. Below, we explain how we designed the system to ensure these properties and how we evaluated the result.

**Phrase 19**

| Segment 1 | Segment 2 | Segment 3 | Segment 4 | Segment 5 |
|---|---|---|---|---|
| an allen Expositionen | wachsen | die | Triebschneeansammlungen | an. |
| in allen Gebieten | wuchsen | die bereits grossen | | weiter an. |
| {vor_allem} in Kamm- und Passlagen | | die zuvor kleinen | | nochmals an. |
| {vor_allem} {Gebiet} {,Gebiet} {und_Gebiet} | | | | stark an. |
| lokal | | | | kaum noch an. |
| in den letzten {Anzahl} Tagen | | | | etwas an. |
| mit {dem_Wind} | | | | deutlich an. |
| mit Neuschnee und Sturm | | | | |
| ... (total: 41 options) | | | | |

**{sub-segments}** (for segment 1, option 4)

| {vor_allem} | {Gebiet} | {,Gebiet} | {und_Gebiet} |
|---|---|---|---|
| [Empty] | in allen Gebieten | [Empty] | [Empty] |
| vor allem | am Alpennordhang {{östlich}} {{Ort}} | (-), in der Urseren | und in den östlichen Voralpen |
| besonders | in den oberen Vispertälern | (-), am Alpensüdhang | und im Gotthardgebiet |
| auch | entlang der Grenze zu Italien | (-), in Nord- und Mittelbünden | und am Alpensüdhang |
| | ... (total: 232 options) | ... (total: 233 options) | ... (total: 233 options) |

**{{sub-sub-segments}}** (for sub-segment {Gebiet}, option 3)

| {{östlich}} | {{Ort}} |
|---|---|
| östlich | von Interlaken |
| westlich | der Reuss |

**Fig. 3.** Schema of phrase number 19 in the source language German (for reasons of space we cannot show all the predefined options nor the presentation in the operational editor). The phrase consists of a succession of five segments. For each segment, the authors can select from a pull-down menu of predetermined options, except for segment 4, which contains a fixed text. Some of the options in segment 1 consist of one or multiple sub-segments {in brackets} with selectable options (shown only for option 4). The first option of the sub-segment {Gebiet} contains two sub-sub-segments {{in double brackets}} with selectable options (shown at the bottom). The blue options are an example that make sense in German and must thus be correct in all the target languages as well. The red options mark a combination that didn't make any sense grammatically or with respect to content. Such combinations are not chosen from the forecasters and can be nonsense in the target languages as well. "(-)" is a control character which indicate, that there is no space before. Adjectives ("grossen", "kleinen" in Segment 3 are only allowed, if they refer always and in all the languages to subjects with the same gender and number. In this example, this is obviously fulfilled, as there is only one possible subject, namely "Triebschneeansammlungen" in segment 4).

### 3.1 Creating the phrases in the source language

The sentences were created by an experienced avalanche forecaster whose native language is German and who has a good knowledge of all the target languages. Numerous avalanche bulletins from the past 15 years were consulted in order to cover as many situations as possible. Phrases were not taken over in a verbatim manner, but their content was generalized and the phrase structure was simplified whenever possible. The main challenge was to find sentences that are universal enough to describe all possible danger situations and simple enough to be directly translated. In contrast to other CNL approaches, no explicit simplified grammar was used. As our sentences need to have direct mappings to all target languages, we had to consider the following points when defining the original German sentences:

- Adjectives can only be used if they refer to subjects with the same gender and number for all options and all languages, or if they have only a local scope occurring together with the subject in the same option. In the first case, gender and number of the subjects in all target languages have to be considered when defining the sentence pattern in the source language. To simplify the catalogue, some terms had to be changed, e.g. the Italian word for 'full-depth avalanche' was changed from the masculine 'scivolamento da reptazione' into 'valanga da reptazione' in order to make it feminine like all other types of avalanches and snow slides.
- Articles depend on number and – in most of the languages used – gender, and must therefore usually be included in the same option as the noun.
- Prepositions often change with the noun and must therefore also be included in the same option as the noun, e.g. 'in' Ticino (a region), but 'on the' Rigi (a mountain).
- As German has four grammatical cases, it is sometimes necessary to split certain phrases into additional segments and sub-segments. E.g. "Fresh snow drifts require caution / are to be avoided" must be separated from "Fresh snow drifts represent the main danger", because in German the case of "fresh snow drifts" turns "Frische*n* Triebschnee beachten / umgehen" into "Frische*r* Triebschnee ist die Hauptgefahr".
- Pronouns are only allowed to refer to nouns in the same sentence. An exception of this restriction are demonstrative pronouns, which can also refer to the immediately preceding sentence, but are only allowed to substitute one specific noun. Thus, e.g. the German "diese" (meaning "these") is listed twice in the same pull-down menu, once for "the avalanches" (in Italian the feminine "queste ultime") and once for "the snow drifts" (in Italian the masculine "questi ultimi"). As there is no difference in the source language German, the substituted noun is indicated in the bulletin editor beside the pronoun, which allows the avalanche forecasters to choose the correct option. This proved to be easy when creating a new sentence but caused mistakes in isolated cases when a sentence was copied from an old bulletin.

Our catalogue of phrases attempts to provide only a single possibility to describe the same content in order to minimize the size of the catalogue and to make it easier to find the required sentence patterns. Exceptions to this rule were made in cases with common German synonyms, where both terms were listed (e.g. "Wiesenhänge" and

"Grashänge" for grassy slopes). This allows forecasters to find the matching sentence, regardless for which of the synonyms they are searching.

When building the catalogue, special attention was paid to extreme situations. We individually checked all situations with "very high" avalanche danger (level 5) from the last 15 years to ensure that our catalogue can accurately manage these extreme cases.

Finally, we get 110 individual sentence templates, which consist of a succession of up to ten segments (mean: 4.3). Each segment consists of 1 to more than 250 predetermined options, which often consist of a series of sub-segments and, as part of the sub-segments, sometimes even sub-sub-segments. All in all we get 603 different lists of selectable options. The individual lists are used in one to 34 different sentence templates, and within a sentence template from once up to several hundred times. The system has the potential to generate a total of several trillion different sentences.

### 3.2  Translation of the catalogue

In our system, translations take place on the segment level. Although German, French, Italian and English are all Indo-European languages, the differences in their grammatical structure including word order, gender, and declension make segmented translation difficult. Thus, specific editing and visualization software had to be developed by a translation agency to prepare the phrase translations. The translations of the segments were then performed manually by professional translators familiar with the topic and applying our text corpus. In addition to the omnipresent problem of inflection, ensuring the correct word order also proved difficult. Other problems included:

- clitics, apostrophes and elisions to avoid hiatus, especially in French and Italian;
- the Italian impure 's' ("*i grandi* accumuli" but "*gli spessi* accumuli");
- the split negation in French ("ne ... pas").

Our system uses no logical functions, distinction of cases or post processing when translating the sentences, except for a check to ensure the presence of a space between the different segments and a capital letter at the beginning of each sentence. In comparison to the source language, only two changes were allowed in the target languages on the segment level (Fig. 4, 5): (1) the segment order could vary between the languages (but is fixed for any given language and thus independent from the chosen options) and (2) each segment could be split in two (into ...a, ...b in Fig. 4-6), which is widely used, mainly to construct idiomatic word orders.

Segment splitting is only used in the target languages but not in German. This is necessary because the user interface for the avalanche forecasters currently allows for segment splitting only in the target languages, which is the only reason why the current system is restricted to translations from German and cannot be used in the other directions (e.g. with French as the source language). It would be a challenge to deal with split segments in the user interface to construct sentences, but otherwise our system and our catalogue could in principle also be applied for the other translation directions.

## German

*Nasse Lawinen können an sehr steilen Sonnenhängen gefährlich gross werden.*

| Segment 1 | Segment 2 | Segment 3 | | Segment 4 | Segment 5 |
|---|---|---|---|---|---|
| die Lawinen | **können** | | | | mittlere Grösse erreichen. |
| **nasse Lawinen** | | auch | | oft | **{ziemlich} gross werden.** |
| Gleitschneelawinen | | in diesen Gebieten | | vereinzelt | {wie_weit} vorstossen. |
| | | **{an_steilen} Sonnenhängen** | | weiterhin | |

| {an_steilen} |
|---|
| an |
| **an sehr steilen** |
| an felsdurchsetzten |
| an wenig befahrenen, eher schneearmen |

| {ziemlich} |
|---|
| |
| ziemlich |
| aussergewöhnlich |
| **gefährlich** |

## English

*On very steep sunny slopes wet avalanches can reach dangerously large size.*

| Segment 3a | Segment 1 | Segment 2 | Segment 3b | Segment 4 | Segment 5 |
|---|---|---|---|---|---|
| | the avalanches | **can** | | | reach medium size |
| auch | **wet avalanches** | | also | in many cases | **reach {ziemlich} large size.** |
| in these regions | full depth avalanches | | | in isolated cases | reach {wie_weit}. |
| **{an_steilen} sunny slopes** | | | | as before | |

| {an_steilen} |
|---|
| on |
| **on very steep** |
| on rocky |
| on little-used, rather ligtly snow-covered |

| {ziemlich} |
|---|
| |
| fairly |
| exceptionally |
| **dangerously** |

## French

*Des avalanches mouillées peuvent devenir dangereusement grandes sur les pentes très raides au soleil.*

| Segment 1 | Segment 2 | Segment 4 | Segment 5 | Segment 3 |
|---|---|---|---|---|
| les avalanches | **peuvent** | | atteindre une taille moyenne | . |
| **des avalanches mouillées** | | souvent | **devenir {ziemlich} grandes** | aussi. |
| des avalanches de glissement | | de manière isolée | avancer {wie_weit} | dans ces régions. |
| | | toujours | | **{an_steilen} au soleil.** |

| {ziemlich} |
|---|
| |
| assez |
| exceptionnellement |
| **dangereusement** |

| {an_steilen} |
|---|
| sur les pentes |
| **sur les pentes très raides** |
| sur les pentes rocheuses |
| sur les pentes peu féquentées, plutôt faiblement enneigées |

## Italian

*Sui pendii soleggiati molto ripidi, le valanghe bagnate possono raggiungere dimensioni pericolosamente grandi.*

| Segment 3 | Segment 1 | Segment 2 | Segment 4 | Segment 5 |
|---|---|---|---|---|
| | le valanghe | **possono** | | raggiungere dimensioni medie. |
| anche | **le valanghe bagnate** | | spesso | **raggiungere dimensioni {ziemlich} grandi.** |
| in queste regioni, | le valanghe da reptazione | | a livello isolato | avanzare {wie_weit}. |
| **sui pendii soleggiati {an_steilen},** | | | ancora | |

| {an_steilen} |
|---|
| **molto ripidi** |
| rocciosi |
| poco frequentati e scarsamente innevati |

| {ziemlich} |
|---|
| piuttosto |
| eccezionalmente |
| **pericolosamente** |

**Fig. 4.** Schema of phrase number 65 in the source language German and the target languages English, French and Italian. In reality, there is an additional segment describing the starting zone and much longer lists of predefined options. {Curly brackets} mark sub-segments: for the two that are chosen, some of the predefined options are listed. [blank] is one of the options in the third and fourth segment. The segments appear in a different order depending on the language. In English, segment 3 is split. The French word for slopes ("pentes") is included in the sub-segment {an_steilen}, while it is part of the main option in the other languages.

**Fig. 5.** Schema of phrase number 57 in the source language German and the target languages English, French and Italian. In reality, there is an additional segment describing the avalanche size and much longer lists of predefined options. The segment order is different in German as compared to the target languages. Segment 1 contains the demonstrative pronoun "They" two times, because of a difference in Italian. In the source language, it is indicated for what the pronoun is staying (2nd option: "avalanches", 4th option: "snow drift accumulations"). The French word for "released" ("déclenchées") is included in segment 3 to avoid a split for the 3rd option. The Italian word for "can" ("possono") is moved to segment 2a to avoid a further split of segment 2 caused by the negation of the 3rd option. The combination "Snow drift accumulations can scarcely be released, even by a single winter sport participant" makes no sense in regard to content. As this is the case in the source language German too, forecasters never use this combination to describe the actual avalanche danger.

Apostrophes, elisions, clitics and the impure 's' are handled by using pull-down splits or by putting the respective parts together into the same option. The latter sometimes require splits across the constituent units. As splits are invisible, this does not reduce the quality of the output (Fig. 6).

For sentences that describe a process in the future, the author of the catalogue gives the estimation of when this will happen. This was needed mainly for the English translation and enables our system to choose the correct future tense.

| Phrase 22 | | | | | |
|---|---|---|---|---|---|
| Segment 1 | Segment 2a | Segment 3 | Segment 4 | Segment 2b | Segment 5 |
| il legame | de(-) | gli | accumuli di neve ventata | [Empty] | è in corso. |
| | reciproco de(-) | i vari | | [Empty] | è insufficiente. |
| | con la neve vecchia de(-) | | | [Empty] | è sfavorevole. |
| | tra i vari accumuli di neve ventata e quello tra | | | e la neve vecchia | è gia piuttosto buono. |
| | | | | | ... (total: 10 options) |

**Fig. 6.** Phrase 22 in Italian, with a split of segment 2. "(-)" is a control character which indicates that there is no trailing space. The construction allows e.g. the following sentences: "Il legame **degli** accumuli di neve ventata è …", "Il legame **dei vari** accumuli di neve ventata è …", "Il legame tra i vari accumuli di neve ventata e quello **tra gli** accumuli di neve ventata e la neve vecchia è …"

### 3.3   Testing the output of the translation catalogue

Not all possible sentences are meaningful (see e.g. Fig. 3, 5), but all those that make sense content-wise must be correct in all languages. As no proofreading is possible in operational use, the translations in the catalogue must be guaranteed to be of high quality, even though the trillions of possibilities cannot be checked individually. Special attention had to be paid to the optional segments, where [blank] can also be chosen. Three different quality checks were conducted for each phrase:

- A sentence generator was created to randomly generate a number of permutations from a phrase. The generated sentences were individually checked by the translators.
- All the options of every segment, sub-segment and sub-sub-segment were checked sequentially by the author of the phrase catalogue in all four languages together. This involved checking both the correctness of the phrases and the content of the translations. In this check, the translation of the catalogue proved to be even more exact with regard to content than the manual translations of the old avalanche bulletins.
- Fictitious avalanche bulletins were written and checked by native-speaker avalanche experts. The operational technical system including the newly developed bulletin editor was also tested in this procedure.

### 3.4   Operational use

Since going operational, about 2000 danger descriptions have been produced per language. As before, the danger descriptions in German were proofread and discussed by at least two avalanche forecasters. Once the content of the German text was found to be correct, about 150 different products (online, app and print products including regional danger maps - they all contain the same danger descriptions) were generated for each bulletin edition and published automatically [16]. This includes publishing the automatically translated texts without any further proofreading or corrections. The new system reduces the time span between the last possible change in the danger description (in the source language German) and the time of publishing it in all four languages from about one hour to a few minutes.

The automated generation makes the avalanche bulletin highly dependent on the technical systems. Fortunately, the catalogue of phrases proved to be very stable: in the first two seasons of operational use, only 1 of approximately 8000 danger descriptions was not made available to the public within 2 hours due to technical problems with the catalogue. This leads to a technical availability of 99.999% for the catalogue alone, which exceed the reliability of most of the other systems needed for the avalanche bulletin as e.g. the web servers.

The use of the catalogue did not change the working load of the avalanche service in the last two winters. Obviously, the additional time needed to search the matching sentences was compensated by the savings due to the fact that double-checking the translated texts is no longer necessary and due to the new editor.

## 4　Evaluation of the accuracy with regard to content

### 4.1　Method

To analyze the usability of the catalogue of phrases, we first had to measure the precision with which the avalanche forecasters can describe the different avalanche situations under operational conditions.

We considered several possible methods to evaluate the accuracy of our system. We realized that a comparison of the real avalanche situation, if it were known, with the description in the bulletin would be more a measure of the prediction quality than of the possibilities offered by the catalogue of phrases. Also rewriting old avalanche bulletins is not sufficient, because it leaves out the fact that appropriate sentences not only have to be present but they also have to be found by the avalanche forecasters in charge within reasonable time. Writing an avalanche bulletin under operational conditions always includes stress and time pressure and cannot be simulated in a scientific study. Furthermore, in old danger descriptions it always rests somewhat unclear what part of the content was chosen very consciously and must thus be preserved exactly, and which parts were less certain and could have been stated in a different way.

We came to the conclusion that the most accurate way to evaluate our catalogue of phrases is to give a simple questionnaire to the avalanche forecasters asking to what degree accuracy they could formulate what they intended to describe. This was done in a survey in February 2014 (Fig. 7).

The catalogue of phrases is currently used by Switzerland's seven avalanche forecasters. They all filled out our questionnaire, but the number of participants is obviously still too small for sophisticated statistical analysis. Nonetheless, the answers aggregate the cumulative experience of the about 2000 danger descriptions that were written with the catalogue system up to now. In the case of missing sentences, the system would allow users to add arbitrary text strings in all four languages and to use them immediately. However, no such 'joker phrases' were actually used during the first two winters of operational service.

## 4.2 Results

To write exactly what they wanted was possible "almost always" for five forecasters and "in most of the situations" for the other two forecasters. The differences in what the forecasters wanted to write and in what they could write with the catalogue where often so small, that six out of seven forecasters declared that "almost always" the differences where within the uncertainty in terms of knowledge of the actual situation. Four forecasters stated that the users would never notice a difference in what was the original idea of writing and the text effectively written with the catalogue of phrases, while the others responded that "almost always" the users would not notice a difference.

On the other hand two forecasters from time to time had "minor, but noticeable limitations", while this occurred to the other five forecasters only "very seldom". "Major limitations" or imprecise descriptions never occurred.

All the avalanche forecasters rated their satisfaction with the catalogue with the best grade ("excellent"). Our decision to build a system based on a catalogue of phrases for the avalanche bulletin was for one forecaster "good" and for the others even "the real deal". Asked for the major advantages of the catalogue of phrases, the forecasters named:

- Immediate high quality translation in multiple languages.
- No proofread nor corrections needed.
- Standardised formulations.

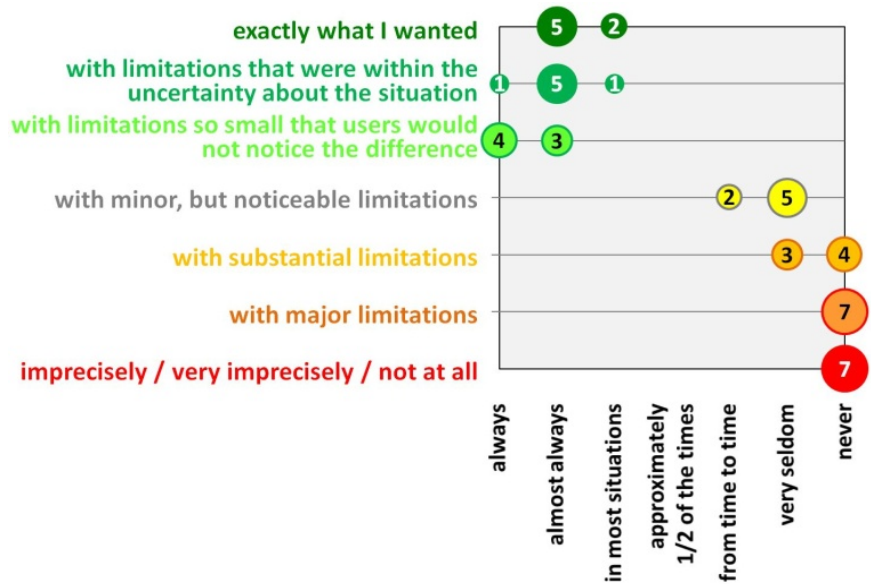

**Fig. 7.** Answers given by the seven Swiss avalanche forecasters on the question: *"How accurately could you describe the situation?"*

As weak points, the forecasters saw mainly:

- The difficulty to find the right sentences under time pressure.
- The intricate maintenance to keep the catalogue up-to-date over a long time period (Fortunately, this was not really a problem until now. One minor update of the catalogue was done after the first season of operational use, and the second is planned for autumn 2015).
- The dependency on a sophisticated technical system as well as on the author of the catalogue.

## 5      Evaluation of the quality of the texts

### 5.1    Method

How can we measure the language quality of an avalanche warning? Language quality, as we use the term here, includes the aspects of grammatical correctness, clearness of expression and understandability. With respect to the content, we simply asked the forecasters via a questionnaire. In regard to the language quality, we could have asked professional linguists, but there would still remain the question of whether it really is a language that is spoken and understood by the customers of avalanche warnings, that is backcountry tourers and freeriders. For this reason, we instead decided to directly ask the users of the avalanche bulletin and to analyze their ratings statistically.

As language quality is a quite subjective issue, we chose a comparative approach, letting users rate manually written texts as well as texts from our catalogue in two different surveys.

### 5.2    Setup of survey 1

The main objectives of the first survey were to monitor user opinions on the avalanche bulletin in general and to evaluate the changes due to the new avalanche bulletin [20]. In this survey we also placed two questions about the language quality (Tab. 3).

**Table 3.** *"How do you evaluate the avalanche bulletin as regards the following criteria?"*

|  | very good | rather good | neither good nor bad | rather bad | very bad |
|---|---|---|---|---|---|
| quality of articulation: description of dangers |  |  |  |  |  |
| quality of articulation: section "snowpack and weather" |  |  |  |  |  |

The survey was posted on www.slf.ch, on the website of the Swiss avalanche warning service, from 18 February to 5 March 2014. The questions about the language quality were not mandatory, and after filtering out participants who obviously did not take the survey seriously (e.g. mountain guides younger than 20 years or without experience, or participants who always marked the top grade) we ended up with 2313 participants who answered both questions. 83% of them were men and the mean age was 42 years. The number of answers per language reflects well the proportion of visits on our website (Tab. 4), except for English participants, which were underrepresented in the survey ($p<0.001$).

Table 4. Answers per language in survey 1

|  | German | English | French | Italian |
| --- | --- | --- | --- | --- |
| answers in survey 1, ratio | 1673, 0.72 | 50, 0.02 | 428, 0.19 | 162, 0.07 |
| ratio on website | 0.71 | 0.04 | 0.19 | 0.07 |

### 5.3 Setup of survey 2

In the second survey, we compared texts from old, manually written and translated danger descriptions with new danger descriptions from the catalogue. This survey was a blind study and only contained questions with respect to the language quality.

To get a comparable set, we chose one danger description from the evening edition of the avalanche bulletin from every second calendar day, starting at the beginning of December and finishing at the end of March. The descriptions from the catalogue were taken from the 2012/13 winter season, the freely written descriptions from the issues from winter 2011/12 back to 2007/08. To avoid evaluating texts that were too short, we only used danger descriptions with more than 100 characters in German. On days with more than one danger description, we randomly chose one of them.

As 120 descriptions per language are too much for a survey, we divided them into 6 different sets, containing 10 descriptions from the old and new bulletin each. We divided the descriptions into the different sets in such a way that different avalanche situations were distributed as uniformly as possible. The order of the descriptions was chosen randomly for each dataset, but was identical across all languages.

For every description, we asked four questions about the language quality (Tab. 5) and, additionally, in what manner the participant thought the text was produced (manual versus automatic).

The survey was posted on www.slf.ch, on the website of the Swiss avalanche warning service, from 18 February to 5 March 2014, together with survey 1. Each participant randomly received one out of the six data sets, in the same language as the website was visited. After filtering out spam and otherwise non-serious replies, we ended up with data from 204 participants (Tab. 6).

Of the 198 participants who answered the question of whether they are native speakers, 95% in fact were (German: 97%, English: 59%, French and Italian: 100%). 81% of the participants were men, and the mean age was 43 years. Reflecting the languages of the visitors of our website, we received most answers for German (76) and the least for English (18). We also asked about the participants' own prior experience in evaluating avalanche dangers, with the result that English participants rated their own experience slightly higher (median between "medium" and "high") than the other participants ("medium").

**Table 5.** Questions concerning the language quality (correctness, comprehensibility, readability and clarity)

| 1. *"Is the text correct?"* ||||
|---|---|---|---|---|
| ("minor error" = typing mistake, incorrect punctuation or use of upper/lower case letters...) |||||
| Absolutely correct | 1 minor error | several minor / 1 major error | several major errors | Completely wrong |
| 2. *"Is the language easy to understand?"* |||||
| (Assuming familiarity with the key technical terms) |||||
| Very easy to understand | Easy to understand | Understandable | Difficult to understand | Incomprehensible |
| 3. *"Is the text well formulated and pleasant to read?"* |||||
| Very well crafted | Easy to read | Clear | Difficult to read | Barely or not at all readable |
| 4. *"Is the situation described clearly?"* |||||
| Clearly and precisely | Reasonably clearly | Understandably | Unclearly, meaningless | Incomprehensibly, contradictory |

**Table 6.** Participants in the survey, divided into languages and allotted datasets

|  | German | English | French | Italian |
|---|---|---|---|---|
| participants | 76 | 18 | 55 | 55 |
| per set (1/.../6) | 14/11/13/10/16/12 | 3/2/3/5/4/1 | 10/12/9/6/6/12 | 9/9/7/12/8/10 |

### 5.4 Analysis

To analyze the detection rate of the origin (i.e. manual or automatic creation) of a description within a language, the data were cross-tabulated and the chi-square statistic was calculated. All differences between categorical variables were tested with the Mann-Whitney U-test for statistical significance (using $p=0.05$).

When comparing the language quality from freely written texts to those from the catalogue of phrases, we could only find differences in isolated cases by using common parameters for ordinal data as median or mode. As we did not wish to jump to the conclusion that there was no difference at all, we assumed the predefined responses to be equal in distance and allocated numerical values to the different categories, starting with 5 for the best rating and 1 for the worst. We only used these numerical values to calculate mean values in order to show differences between different languages and between the origins of the descriptions.

In survey 2, not all of the six datasets of a particular language had the same number of usable answers (Tab. 6). We therefore checked our data in every language for anomalies in the distribution between the different datasets. As we did not find any, we pooled all answers together for further analysis.

To evaluate the differences between the languages, as well as to analyze the overall rating over all the languages, we used in survey 2 a balanced dataset. This contained all English answers and for each of the other languages randomly chosen ratings of 180 descriptions from each of the old and the new bulletin.

### 5.5 Results

Table 7 shows the answers of survey 1. The language quality of the danger descriptions (which originate from the catalogue) is rated better than the snowpack and weather section (which is conventionally written and manually translated). This effect can be found in all the languages, but is only significant in German, due to the larger number of participants.

**Table 7.** Rating for the danger descriptions and difference between the danger descriptions and the "snowpack and weather" sections. Better ratings for the danger descriptions are marked green, lower ratings did not occur. Significant differences are highlighted.

|  | German | English | French | Italian |
|---|---|---|---|---|
| mean | 4.47 | 4.28 | 4.36 | 4.45 |
| difference (danger descriptions - snowpack and weather) | **0.07** (p=0.002) | 0.10 (p=0.55) | 0.05 (p=0.18) | 0.04 (p=0.51) |

In survey 2, the evaluators detected the origin of a given text in 59% of the German descriptions (Tab. 8). In the target languages, the rate of correct recognition was lower, with 55% in French and 52% in Italian and English. The recognition rate was significantly better than random only for German and French.

Table 9 shows the answers to the questions regarding the real origin of the danger descriptions. Differences between old and new descriptions are small and vary from language to language. Table 10 shows the text length of the danger descriptions that were used in survey 2. German as source language has the shortest lengths for both kinds of texts.

**Table 8.** Correct guesses with respect to the origin (manual versus automatic creation) of the descriptions. Significant values are highlighted.

|  | German | English | French | Italian |
|---|---|---|---|---|
| n (equally balanced old/new) | 1520 | 360 | 1100 | 1100 |
| detection rate | **0.59** | 0.52 | **0.55** | 0.52 |
| p - value | p < 0.001 | p = 0.40 | p < 0.001 | p = 0.13 |

**Table 9.** Ratings for the new descriptions from the catalogue of phrases and the differences between new and old descriptions. The column header refers to the questions shown in Tab. 5. Better ratings for the new descriptions are marked green, lower ratings red. Significant differences are highlighted. *are calculated from the balanced dataset.

|  |  | correct | comprehensible | readable | clear | all |
|---|---|---|---|---|---|---|
| German (n=1520) | mean | 4.75 | 4.30 | 3.93 | 4.29 | 4.32 |
|  | difference (new-old) | 0.03 (p=0.22) | **0.13** (p=0.003) | 0.05 (p=0.25) | **0.16** (p=0.001) | **0.09** (p<0.001) |
| English (n=360) | mean | 3.89 | 3.74 | 3.51 | 3.73 | 3.72 |
|  | difference (new-old) | -0.01 (p=0.61) | 0.01 (p=0.90) | 0.03 (p=0.95) | -0.05 (p=0.45) | -0.003 (p=0.54) |
| French (n=1100) | mean | 4.57 | 4.30 | 4.07 | 4.34 | 4.32 |
|  | difference (new-old) | **-0.12** (p=0.001) | -0.04 (p=0.42) | **-0.11** (p=0.01) | 0.01 (p=0.47) | **-0.07** (p=0.001) |
| Italian (n=1100) | Mean | 4.35 | 4.21 | 3.99 | 4.28 | 4.21 |
|  | difference (new-old) | **-0.16** (p=0.001) | -0.09 (p=0.08) | -0.08 (p=0.12) | **-0.12** (p=0.01) | **-0.11** (p<0.001) |
| all languages | mean | 4.39 | 4.14 | 3.87 | 4.16 | 4.14 |
|  | difference (new-old) | -0.06 (p=0.08)* | 0.004 (p=0.21)* | -0.03 (p=0.10)* | 0.001 (p=0.08)* | **-0.02** (p=0.02)* |

**Table 10.** Text length of the evaluated danger descriptions

|  | characters (mean value) | length in relation to German (mean value) | | |
|---|---|---|---|---|
|  | German | English | French | Italian |
| conventionally written and translated | 288 | 121% | 132% | 125% |
| catalogue of phrases | 319 | 119% | 125% | 125% |

## 6 Discussion

Our accuracy evaluation showed that the catalogue of phrases has always allowed an adequate description of the danger situation by the avalanche forecasters. The translations in the catalogue were checked extensively by the developer, an experienced avalanche forecaster with knowledge in all four languages. The catalogue proved to be even more exact with regard to content, probably because of the fact that the manual translation method used for old avalanche bulletins lacked the necessary time to correct smaller inconsistencies.

The evaluations on the language quality showed that the detection rate of the origin of the danger descriptions was barely better than random, with an average of 55% correctly recognized descriptions for all languages. To put this into perspective, this corresponds to, for example, correctly recognizing 2 out of 20 descriptions and then tossing a coin for the remaining 18 descriptions.

Texts from the catalogue are rated very similarly as compared to the conventionally written and manually translated texts, both in source and target languages. The following differences were observed among languages and surveys:

- In survey 1, the danger descriptions (originating from the catalogue) were rated better than the text in the snowpack and weather section (which is still manually written and translated). However, the difference was only significant in German, the source language, which also had most ratings.
- In survey 2, the differences vary from language to language. German speaking participants rated the texts from the catalogue significantly better (for two of the criteria and in the mean over all criteria), whereas French and Italian participants rated the new texts significantly worse (also in two criteria and in the mean over all criteria). In English we could not find any such differences. Next, we calculated the overall ratings for survey 2 with a balanced dataset, and we only get a significant decrease when taking all languages and all questions together (p=0.02), but not for individual questions. With the more than 5,500 assessments we get a statistically significant result, even though the decrease is marginal with a value falling from 4.16 to 4.14.

The introduction of the catalogue of phrases was a fundamental change and the catalogue itself was mostly translated by different translators. It is therefore surprising that there is only a marginal difference between manually written texts and those originating from the catalogue. The differences between the languages were however much larger and more consistent between the two surveys:

- The language quality of the texts from the catalogue is rated best in the source language German with an overall value of 4.40 (4.47 in survey 1, 4.32 in survey 2), but also the ratings for French and Italian were appealing with overall values of 4.34 (4.36; 4.32) and 4.33 (4.45; 4.21) respectively.
- The English overall rating was lower with 4.00 (4.28, 3.72). Perhaps this is due to the fact that our translators are British and we use the glossary of the European Avalanche Warning Services (www.avalanches.org), which differs substantially from terminology used in North America, where at least some of the participants of the survey live. As a consequence of this rating, the forecasters are now controlling the English technical terms used in the Swiss avalanche bulletin.

The large variance between different participants in assessing the same dataset in survey 2 shows, that the absolute value of the rating is not only a question of the texts, but also affected by varying interpretations. To understand this effect, further research would be needed. In our survey, we are much more interested in the changes in language quality due to the introduction of the catalogue of phrases than in the absolute value of quality.

Of all properties in survey 2, the correctness reaches the best rating (Tab. 9), whereas the comprehensibility and the clarity of the formulations lie ex aequo in the middle of the investigated parameters about language quality. The catalogue of phrases leads to a standardized language. As Swiss avalanche forecasters believe that this kind of simple and unambiguous language is well suited to communicate warnings, they wrote the "old" danger descriptions in a similar style. In this context, it is not surprising that of all the quality criteria, the pleasure to read was assessed lowest. Note that some users also commented in survey 1 that the text of the new bulletin is somewhat boring after reading it two times every day and the situation does not change for a certain time. That is due to the inevitable monotony of a controlled language in which only one sentence exists to express a given content.

Warnings should in general be as short as possible. In German, the mean length of the danger descriptions from survey 2 is 288 characters if manually written and 11% longer with the catalogue of phrases (Tab. 10). With the new avalanche bulletin, some limitations in the text lengths have fallen away and the forecasters sometimes use this to give more information than what was possible before. Thus, it remains unclear whether the catalogue of phrases really leads to longer texts for the same content. Even if it does, the increase is at most 11%. German as source language has the shortest text, independent from the mode of creation. English texts are about 20% and Italian texts 25% longer than in German, independent from the mode of creation. French texts were 32% longer than in German if written manually. This additional length decreased significantly with the introduction of the catalogue to 25% ($p=0.008$).

Given the poor recognition rate as well as the only marginal and contradictory differences between texts that are conventionally written and texts originating from the catalogue of phrases, we conclude that in the majority of cases, users did not notice any change in the language quality with the introduction of the catalogue. Thus, the language quality from the catalogue of phrases can be judged as virtually equivalent to the texts written from scratch and translated by topic-familiar professionals.

## 7 Conclusions

The catalogue-based system proved to be well-suited to generate the Swiss avalanche bulletin. After two years of operational use, all forecasters declared that within the limited time available to produce forecasts, it was possible to describe the different avalanche situations with high precision and efficiency.

The system also proved to be well-suited for fully automatic and instantaneous translation of danger descriptions from German into the target languages French, Italian and English. The translations do not need to be proofread or corrected, and they turned out to be even superior with respect to their content than the manual translations of the old avalanche bulletin.

The quality of the language was assessed in two different user surveys by comparing conventionally written and manually translated texts to the texts originating from the catalogue of phrases. To get reliable results, we only used texts from real avalanche bulletins.

Recognizing the origin (i.e. created manually or by the catalogue system) of the danger descriptions proved to be difficult; the mean detection rate was only 55%.

Based on different criteria, the quality of the texts was rated by the users as being good, with some differences between languages. With the introduction of the catalogue of phrases, there were only marginal changes in the different quality ratings. Depending on the language and the survey, our system showed a small improvement or a slight decrease in quality. Thus the texts produced by the catalogue of phrases were virtually equivalent in language quality to those produced using the old method of manual translation. The benefit of having immediate automatic translations does therefore not have to be paid with a loss in language quality.

As using a phrase catalogue requires experience, frequent operational use is necessary. It is crucial that users find the phrases matching the given danger situation quickly enough, which has shown to be the case for our system. The implemented search engine was essential. Our experience has shown that the number of phrases should be kept to a minimum by reusing individual phrases in multiple contexts, and that the presented approach is particularly well-suited if the problem domain can be described by a small sublanguage, as is the case for the highly specific topic of avalanche forecasting.

With respect to financial concerns, the cost-benefit ratio of our system turned out to be excellent. The savings from not needing manual translations are expected to exceed the initial development costs within a few years. We believe that such a system could probably be successfully applied in other multilingual countries and could be extended to topics such as weather forecasting. Adapting it to translations for languages from different language families, however, seems difficult due to our simple phrase-based approach that is largely grammar-agnostic.

The construction of the catalogue and the translations had both been based on empirical data. We gladly make them available to interested third parties for further investigations.

## 8 Acknowledgments

We thank Martin Bächtold from www.ttn.ch and his translators for the courage to translate the catalogue of phrases and for maintaining such a high quality standard. Furthermore, we thank Martin Volk, Eva Knop, Nico Grubert, Frank Techel, Chris Pielmeier, Curtis Gautschi and the three anonymous reviewers for valuable input. Finally, we would like to thank the participants of the survey.